\documentclass[runningheads]{llncs}

\usepackage{eccv}

\usepackage{xcolor}
\definecolor{mildgreen}{RGB}{0,120,0}
\usepackage{makecell}

\usepackage{eccvabbrv}

\usepackage{graphicx}
\usepackage{booktabs}

\usepackage[accsupp]{axessibility}  

\usepackage{hyperref}

\usepackage{orcidlink}

\begin{document}

\title{Automated Erythrocyte Detection and Tracking for Retinal Blood Flow Quantification in Erythrocyte-Mediated Angiography} 

\titlerunning{Automated Erythrocyte Detection and Tracking}

\author{Chiao-Yi Wang\inst{1}\orcidlink{0000-0002-5481-8911} \and
Havish S Gadde\inst{2} \and
Yi-Ting Shen\inst{3}\orcidlink{0000-0002-1167-5535} \and
Saige M. Oechsli\inst{2} \and
Osamah Saeedi\inst{2} \and
Yang Tao\inst{1}}

\authorrunning{C. Y. ~Wang et al.}

\institute{Department of Bioengineering, University of Maryland, College Park, MD 20742, USA
\email{\{cyiwang, ytao\}@umd.edu}\\ \and
Department of Ophthalmology and Visual Sciences, University of Maryland School of Medicine, Baltimore, MD 21201, USA \and
Department of Electrical and Computer Engineering, University of Maryland, College Park, MD 20742, USA\\
}

\maketitle

\begin{abstract}
Capillary-level retinal blood flow (RBF) has strong potential as a biomarker for various ocular diseases. However, modalities for measuring capillary-level RBF remain limited. Erythrocyte-mediated angiography (EMA), an emerging imaging technique, enables capillary-level RBF measurement by visualizing individual erythrocytes, yet automated erythrocyte detection and tracking, which are essential for quantifying blood flow, remain largely unexplored.
To address this gap, we propose \emph{EMTrack}, a novel framework featuring a flow-context module for erythrocyte detection that distinguishes moving from paused cells and a topology-aware tracking strategy that enables tracking under large inter-frame displacements and substantial motion variations. In addition, we establish \emph{RBF-EMA}, a new EMA dataset with comprehensive erythrocyte detection and tracking annotations.
Experimental results demonstrate that our method outperforms baseline methods both quantitatively and qualitatively on detection and tracking tasks in the RBF-EMA dataset. Moreover, RBF quantification results highlight the strong potential of our framework for automated retinal blood flow measurement. 
\end{abstract}

\section{Introduction}
\label{sec:intro}
Retinal blood flow (RBF) is associated with a wide range of ocular diseases, including glaucoma \cite{nicolela1996ocular}, diabetic retinopathy \cite{patel1992retinal}, and age-related macular degeneration \cite{ciulla1999color}, as well as neurodegenerative diseases such as Alzheimer's dementia \cite{feke2015retinal, berisha2007retinal}. In particular, capillary-level RBF has the potential to serve as a sensitive biomarker for the early detection of ocular pathologies and may facilitate the development and evaluation of novel therapeutic strategies. However, modalities for measuring RBF remain limited by imaging constraints \cite{hwang2023retinal}, and accurate quantification of capillary RBF remains challenging.

Recently, erythrocyte mediated angiography (EMA) \cite{flower2008observation, pircher2017review, arichika2013noninvasive, lee2017face, riva2010ocular, wang2024memo} has emerged as an imaging technique capable of visualizing individual erythrocytes within the retinal capillary circulation across a large field of view. By capturing the motion of fluorescently labeled erythrocytes \emph{in vivo}, EMA provides the potential to estimate erythrocyte velocity and quantify blood flow at the capillary level. However, existing approaches \cite{saeedi2018determination, chen2024plexus} for measuring absolute erythrocyte velocity in EMA rely on manual identification and tracking of individual erythrocytes by human experts, a process that is highly time-consuming and labor-intensive. This reliance on manual analysis substantially limits the scalability, broader adoption, and practical utility of the modality. Therefore, automated erythrocyte tracking methods are essential to fully leverage EMA, enabling efficient and scalable RBF quantification and facilitating its broader impact as a potential clinical biomarker.

\begin{figure}[tb]
  \centering
  \begin{subfigure}{0.32\linewidth}
  \includegraphics[width=\linewidth]{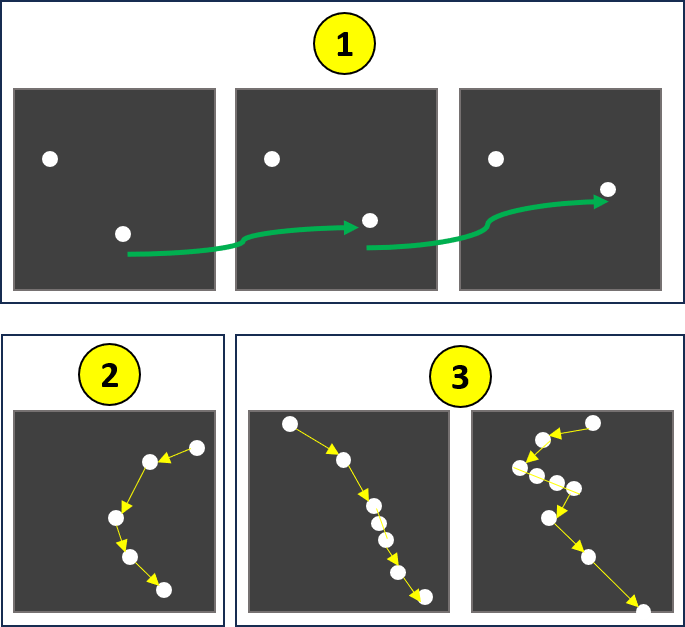}
    \caption{}
    \label{fig:intro1-a}
  \end{subfigure}
  \hfill
  \begin{subfigure}{0.32\linewidth}
    \includegraphics[width=\linewidth]{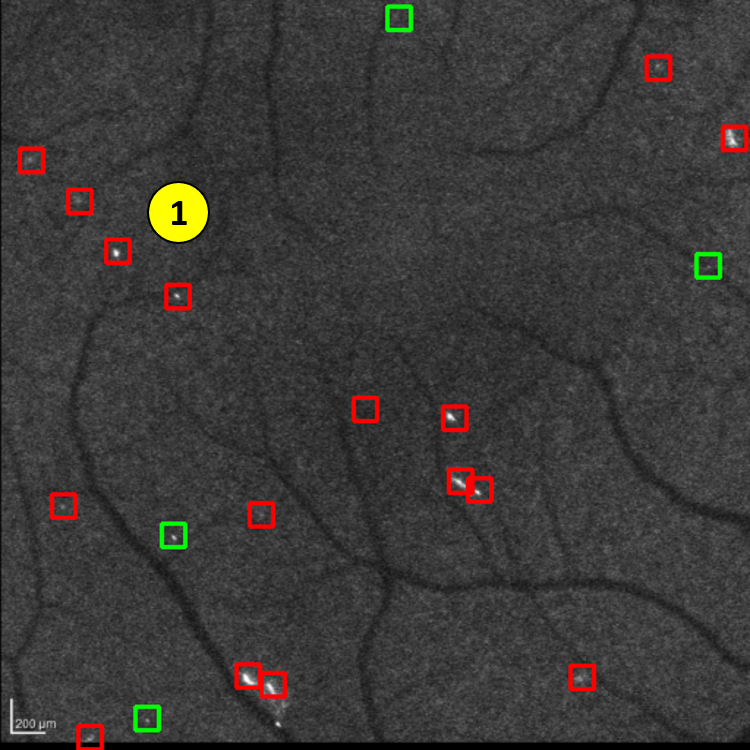}
    \caption{}
    \label{fig:intro1-b}
  \end{subfigure}
  \hfill
  \begin{subfigure}{0.32\linewidth}
    \includegraphics[width=\linewidth]{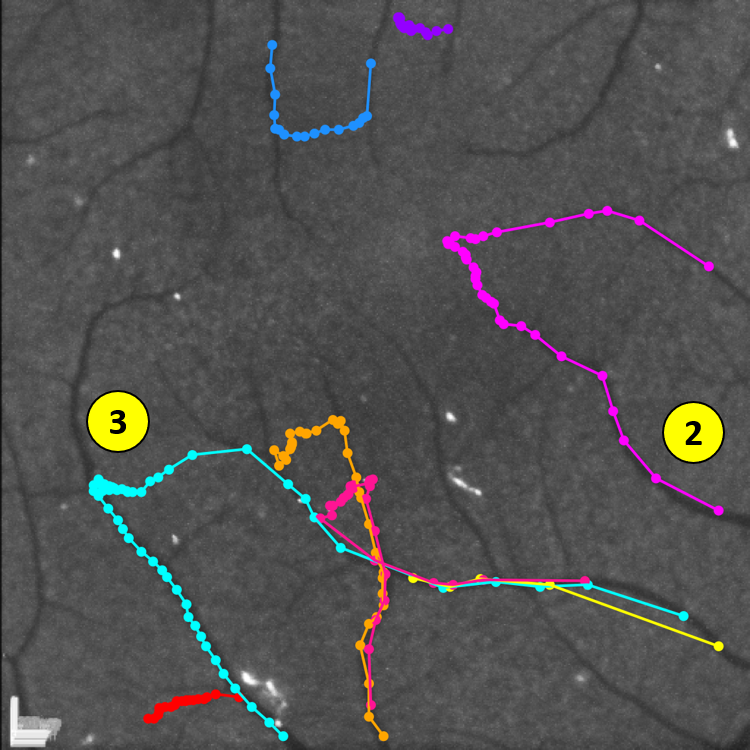}
    \caption{}
    \label{fig:intro1-c}
  \end{subfigure}
  \caption{Challenges in EMA videos for erythrocyte tracking. Unlike typical cell or particle tracking scenarios, EMA videos present three main challenges. (a) Visualization of the three challenges: moving and paused cells exhibit highly similar appearances in individual frames (\textbf{Challenge 1}); large inter-frame displacements (\textbf{Challenge 2}); and substantial motion variations, including abrupt speed changes and sharp turns (\textbf{Challenge 3}). (b) An example EMA image illustrating Challenge 1, where paused cells are marked with red boxes and moving cells are marked with green boxes. (c) Ground-truth trajectories from an example EMA video illustrating Challenge 2 and 3.}
  \label{fig:intro1}
\end{figure}

While automated erythrocyte tracking in EMA shares similarities with both cell tracking and particle tracking, several unique characteristics of EMA images make the task particularly challenging. First, unlike cell tracking in conventional microscopy datasets (\eg, DeepCell \cite{marks2025cellsam} and Bacteria \cite{van2018spatially}), subcellular structures in EMA are not clearly visible. As a result, erythrocytes appear as visually similar particles with limited discriminative features, as shown in \cref{fig:intro1-b}. Due to this highly homogeneous appearance, the problem more closely resembles particle tracking than traditional cell tracking. Second, unlike particles in the ISBI particle tracking dataset \cite{chenouard2014objective}, which typically move randomly within a local neighborhood, erythrocytes in EMA travel along vascular structures, with acceleration and deceleration driven by hemodynamic circulation. This motion pattern produces longer trajectories and larger inter-frame displacements, particularly when cells move through arteries and veins, as illustrated in \cref{fig:intro1-c}, making reliable frame-to-frame association more difficult. In addition, the frame rate of EMA is limited to 15 FPS due to scanning speed constraints of the clinical imaging system required to maintain optical resolution, further increasing the difficulty of accurate tracking. Third, in addition to moving erythrocytes, paused cells frequently appear in EMA images, as shown in (\cref{fig:intro1-b}). These paused cells occur when erythrocytes pass through the choroidal microvasculature \cite{li2023visualization}. They exhibit visual appearances highly similar to those of moving erythrocytes and can only be distinguished using temporal information, such as consistency across multiple consecutive frames. In practice, even human experts often need to examine multiple frames to reliably differentiate paused cells from moving ones. Because paused cells often appear in large numbers, treating them as tracking candidates would substantially increase the number of possible trajectories, making reliable erythrocyte tracking significantly more difficult.

To address these challenges, we proposed \emph{EMTrack}, an automated framework for moving erythrocyte detection and tracking in EMA. EMTrack consists of two main modules: a detection module (\emph{FlowContext-DINO}) and a tracking module (\emph{topology-aware tracking}). In the detection module, temporal information from multiple consecutive frames is incorporated to capture motion cues and combined with visual features extracted by DINO \cite{caron2021emerging}, a self-supervised vision transformer, enabling reliable separation of moving erythrocytes from visually similar paused cells. In the tracking module, vascular structures are first recovered to guide topology-aware tracking. Since capillaries are not directly visible in EMA images, we reconstruct vascular paths by aggregating temporal intensity traces produced by moving erythrocytes across the video sequence, enabling recovery of the underlying capillary structures. The recovered capillary map is then combined with vessel segmentation results for larger vessels to obtain a complete vascular topology. This topology is then incorporated into Dijkstra's algorithm \cite{EDijkstraNM1959} to guide data association in the tracker along vascular structures. This design effectively handles large inter-frame displacements and velocity variations caused by hemodynamic circulation.

To evaluate the proposed framework, we collected an EMA video dataset, \emph{RBF-EMA}, from nonhuman primates. Experimental results demonstrate that the EMTrack accurately detects moving erythrocytes using temporal information and outperforms existing baseline methods in tracking performance. Our contributions are summarized as follows:
\begin{itemize}
\item We propose EMTrack, the first automated framework for moving erythrocyte detection and tracking in EMA.
\item We design FlowContext-DINO, a detection module that utilizes temporal information from consecutive frames together with DNIO features to distinguish moving erythrocytes from visually-similar paused cells.
\item We introduce a topology-aware tracking module that uses vascular topology to guide data association, enabling reliable trajectory linking under large velocity variations and substantial inter-frame displacements.
\item We establish the RBF-EMA dataset, which is the first EMA dataset with both detection and tracking annotations. 
\item Experimental results demonstrate that our method outperforms baseline methods in both detection and tracking tasks.
\end{itemize}

\section{Related Works}
\label{sec:relatedworks}
\subsection{Particle Detection and Cell Detection}
Cell detection and particle detection differ primarily in their detection targets. Cell detection typically focuses on segmenting cells with relatively well-defined and distinguishable morphological structures. For example, Marks et al. \cite{marks2025cellsam} proposed a prompt-engineering approach that employs the Segment Anything Model (SAM) for cell segmentation. In contrast, particle detection aims to identify diffraction-limited spots, which usually appear as small, blob-like signals with limited structural information. For instance, Eichenberger et al. \cite{eichenberger2021deepblink} introduced deepBlink, a deep learning–based method for the automatic detection of diffraction-limited spots.

Most existing methods operate on single frames independently, as their objective is to detect every cell or spot within each image. However, in our setting, moving and paused cells exhibit highly similar appearances in individual frames and can only be reliably distinguished by leveraging temporal information across multiple frames. As a result, single-frame detection is insufficient for our task. To address this limitation, we propose a detection module, FlowContext-DINO, which explicitly utilizes consecutive frames to differentiate moving cells from paused ones.

\subsection{Particle tracking and Cell Tracking}
\label{related2}
Particle tracking and cell tracking are closely related problems that have been extensively studied in the literature. In cell tracking, Galluser et al. \cite{gallusser2024trackastra} proposed TRACKASTRA, a general-purpose framework that leverages a transformer architecture to learn pairwise cell associations across frames. However, this method relies on annotated tracking data for supervision. Zhou et al. \cite{zhou2025cellect} introduced CELLECT, a 3D U-Net–based architecture with strong generalization ability, trained using contrastive learning on the Cell Tracking Challenge dataset. In addition, Bragantini et al. \cite{bragantini2024ultrack} proposed Ultrack, which integrates segmentation results from multiple algorithms across frames to construct consistent cell trajectories.

In contrast, particle tracking presents distinct challenges. Unlike cells, particles often lack discriminative appearance features and exhibit little to no spatial overlap across consecutive frames. Consequently, particle tracking methods typically rely more heavily on motion cues and trajectory consistency rather than appearance-based associations. Tinevez et al. \cite{tinevez2017trackmate} proposed TrackMate, a widely used open-source plugin for particle tracking, and Ershov et al. \cite{ershov2022trackmate} later extended it to TrackMate7 by integrating state-of-the-art segmentation algorithms into the tracking pipeline. Zhang et al. \cite{zhang2023motion} further proposed the supervised Motion Transformer Tracker (MoTT), which demonstrates strong performance on the ISBI Particle Tracking Challenge dataset \cite{chenouard2014objective}. However, many existing approaches assume relatively small inter-frame displacement and do not explicitly handle large motion variations. Moreover, some methods depend on dense tracking annotations for training or evaluation \cite{zhang2023motion}, which are labor-intensive and time-consuming to obtain.

\subsection{Blood Flow Quantification}
Automated blood flow quantification methods have been studied in the literature \cite{cheng2025blood, he2023ultra}. However, most existing approaches focus on cerebral imaging \cite{cheng2025blood} or carotid artery imaging \cite{he2023ultra}. For retinal blood flow (RBF) quantification, several imaging modalities other than EMA have been explored, including VISTA OCTA \cite{hwang2023retinal}, AO-OCT \cite{pircher2017review}, AO-SLO \cite{arichika2013noninvasive}, dynamic OCTA \cite{lee2017face}, color Doppler imaging \cite{goebel1995color}, and laser Doppler flowmetry \cite{riva2010ocular}. While these techniques provide valuable insights into retinal hemodynamics, each modality has inherent limitations. For instance, VISTA OCTA, AO-OCT, and AO-SLO typically suffer from a limited field of view, restricting their ability to capture large-scale vascular networks. Laser Doppler flowmetry often exhibits high variability in measured flow rates, which can affect measurement reliability. Additionally, some modalities provide relative rather than absolute velocity measurements, limiting their capacity for precise quantitative analysis.

Compared with these modalities, EMA enables direct quantification of absolute erythrocyte velocity at cellular resolution \cite{saeedi2018determination, chen2024plexus}, offering a unique opportunity for accurate capillary-level RBF measurement. Nevertheless, automated methods for RBF quantification in EMA remain relatively limited. Wang et al. \cite{wang2019automated} proposed a sequential Monte Carlo-based method for automated velocimetry that demonstrated high correlation with ground-truth labeling. However, the approach is not fully automated, as it requires manual pre-selection of target vessels, limiting its scalability and practical applicability.

\section{Method}
\label{sec:methods}
\begin{figure}[tb]
  \centering
  \includegraphics[width=\linewidth]{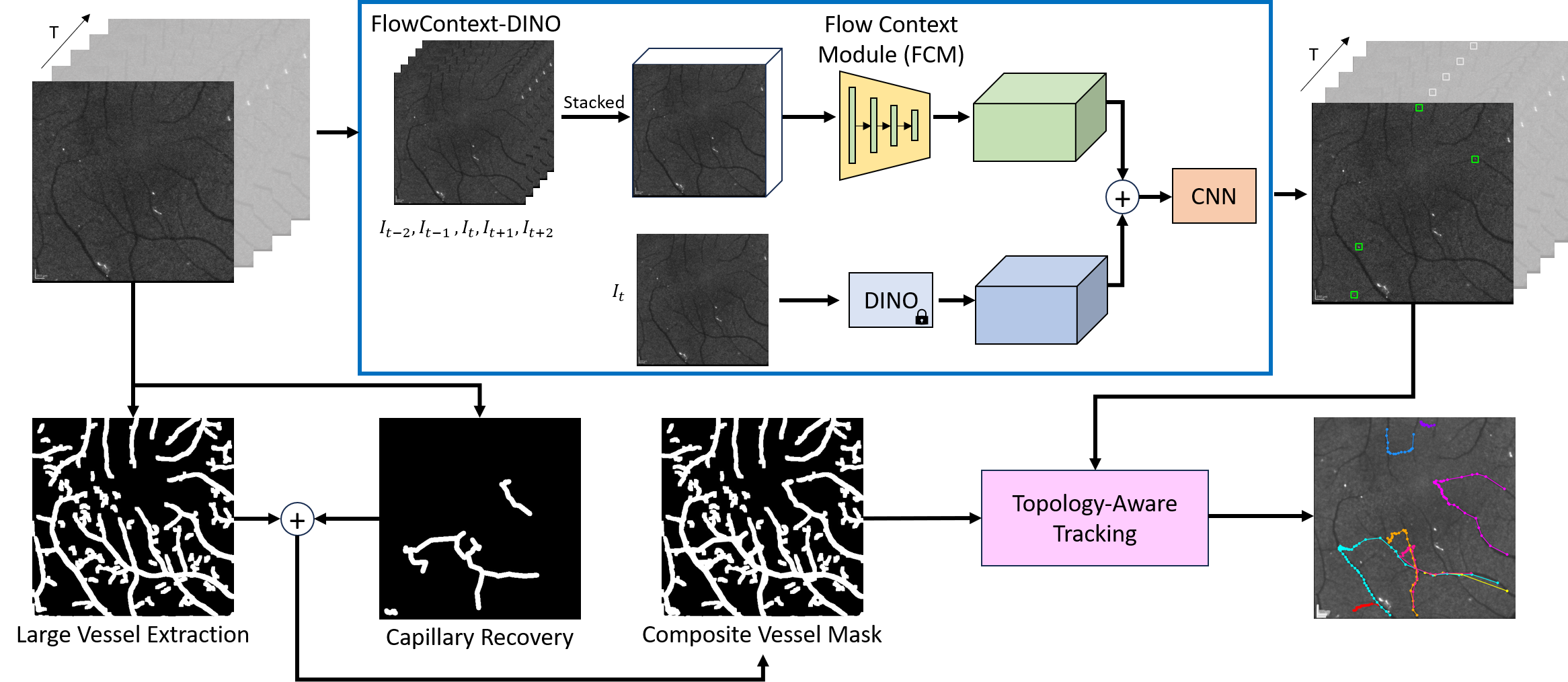}
  \caption{Overview of the proposed EMTrack framework. EMTrack consists of two main components: FlowContext-DINO for moving cell detection and a topology-aware tracking module for cell tracking. FlowContext-DINO enhances a DINO backbone by incorporating a Flow Context Module (FCM) to capture temporal information from $N=5$ consecutive frames. The topology-aware tracking module leverages vessel segmentation masks to improve tracking robustness, particularly in the presence of large inter-frame displacements and substantial motion variations.}
  \label{fig:method1}
\end{figure}

EMTrack (Fig.~\ref{fig:method1}) consists of two components: the FlowContext-DINO module for detecting moving erythrocytes and a topology-aware tracking module. Each component is elaborate below.

\subsection{FlowContext-DINO}

\subsubsection{Feature Extraction.}
Detecting erythrocytes in EMA images is challenging because erythrocytes appear as small particles with limited discriminative visual features and are often affected by imaging noise and intensity variations. Therefore, robust feature representations that capture structural patterns beyond raw pixel intensity are required. DINO \cite{caron2021emerging} is a self-supervised learning framework for training Vision Transformers that produces rich patch-level representations capturing structural and contextual information. These features have been widely utilized in various computer vision tasks, including semantic segmentation \cite{melas2022deep}, depth estimation \cite{zhang2025hybrid}, image retrieval \cite{almazan2022granularity}, and tracking  \cite{tumanyan2024dino}. In this work, we adopt a pre-trained DINO model as the feature extractor.

However, DINO features are extracted independently from individual frames and lack temporal information. In EMA images, moving and paused erythrocytes exhibit highly similar appearances and can only be reliably distinguished using motion cues across multiple frames. Consequently, relying solely on frame-wise DINO features is insufficient for accurately identifying moving erythrocytes.

\subsubsection{Flow Context Module (FCM).}
To incorporate temporal information while preserving the rich representations of DINO features, we design a lightweight Flow Context Module (FCM) that operates on a stack of $N$ consecutive frames centered at the target frame $I_{t}$, while keeping the DINO backbone fixed (\cref{fig:method1}). This design is inspired by prior work \cite{tumanyan2024dino, zhang2023adding}, which trains a small network to predict residual features instead of fine-tuning a large pretrained model. Specifically, the proposed module learns residual feature representations that encode temporal dynamics across frames. This allows the model to incorporate motion cues while preserving the structural and contextual information contained in the original DINO features. The residual features produced by FCM are combined with the DINO features of the target frame to form a refined feature representation:
\begin{equation}
  \Phi(I_t) = \Phi_{DINO}(I_t) + \Phi_{FCM}(I_{stack}),
  \label{eq:method1}
\end{equation}
where $\Phi_{DINO}(I_t)$ denotes the pre-trained DINO features extracted from the target (center) frame $I_t$, and $I_{stack}$ represents a stack of $N$ consecutive frames centered at $I_t$. In our implementation, we set $N=5$.

The refined feature representation is then passed to a single-layer CNN classifier, which produces patch-wise predictions indicating whether each $16\times16$ image patch in the target frame contains a moving erythrocyte. We assume that each patch contains at most one erythrocyte, which empirically holds in our case. For patches predicted as positive, the erythrocyte location $(x, y)$ is estimated by averaging the coordinates of the top-10 brightest pixels within the patch. Finally, non-maximum suppression (NMS) with a distance threshold of 5 pixels is applied to remove overly close detections

\subsection{Topology-Aware Tracking}
As discussed in Section~\ref{related2}, most existing particle tracking methods assume relatively small motion variations and short-range trajectories \cite{chenouard2014objective, tinevez2017trackmate}. In contrast, erythrocytes in EMA travel along vascular structures and can exhibit large inter-frame displacements and substantial velocity variations driven by hemodynamic circulation. In addition, some particle tracking approaches \cite{zhang2023motion} rely on densely annotated trajectories for training, which are labor-intensive and difficult to obtain in EMA datasets. To address these challenges, we propose a label-free topology-aware tracking module that leverages vessel segmentation masks to guide data association along vascular structures, enabling robust tracking under large motion variations.

\subsubsection{Vessel Mask Extraction and Capillary Recovery.}
Due to image resolution limitations, capillaries are not directly visible in EMA images; only arteries and veins can be clearly observed. However, according to hemodynamic circulation principles, erythrocytes travel from arteries through capillaries and subsequently into veins. Therefore, although capillaries are invisible in EMA imaging, their locations can be implicitly inferred from erythrocyte trajectories, particularly in regions where erythrocyte motion is slower within capillary networks. 

Based on this observation, we design a framework to generate a composite vessel mask for each video sequence that incorporates capillary structures. The pseudo-code of the framework is provided in supplementary materials. First, we compute two projection images from the input video sequence: a maximum-intensity projection $I_{max}$, obtained by taking the maximum pixel value over all frames, and an average-intensity projection $I_{avg}$, obtained by averaging pixel values over the entire sequence. The input image for capillary segmentation is then defined as 
$I_{input} = I_{max} - I_{avg}.$
Here, $I_{max}$ preserves signals from all erythrocytes observed throughout the sequence, whereas $I_{avg}$ primarily retains signals from paused cells. Subtracting $I_{avg}$ from $I_{max}$ therefore enhances the contrast of moving erythrocytes. The resulting image $I_{input}$ is subsequently min–max normalized to the range $[0,255]$.

After generating $I_{input}$, a threshold is applied to convert the image into a binary mask. The centroid of each connected region is extracted, retaining one representative pixel per region. Each centroid is treated as a node in a graph. Using KDTree \cite{maneewongvatana1999analysis}, adjacency relationships are constructed by connecting nodes whose pairwise distances fall within a predefined threshold D = 40. The resulting graph is then partitioned into connected components, where each component represents a spatially coherent subset of nodes. Isolated nodes are discarded, and only components with sizes exceeding a predefined threshold are retained. For each retained component, we apply Minimum Spanning Tree (MST) \cite{guttoski2007kruskal} to connect all nodes. The resulting edges are rendered as straight-line segments with a fixed width of 11 pixels, producing the capillary segmentation mask. 

\subsubsection{Composite Vessel Mask.}
To obtain a complete vascular representation, we additionally extract large-vessel structures (\ie, arteries and veins) using the Frangi vesselness filter \cite{frangi1998multiscale}. The resulting large-vessel mask is combined with the capillary segmentation mask to produce the final EMA vessel segmentation mask, which provides the vascular topology used to guide erythrocyte tracking.

\subsubsection{Topology-Aware Tracking.}
We adopt ByteTrack \cite{zhang2022bytetrack} as the tracking backbone due to its simplicity and strong performance in associating detections under low-confidence scenarios. ByteTrack performs data association in two stages, where high-confidence detections are matched first and low-confidence detections are subsequently considered to recover potential missed matches. This design is beneficial in EMA imaging, where erythrocyte detections may have unstable confidence scores due to imaging noise and appearance ambiguity. However, its motion modeling relies on a Kalman filter to predict object locations in subsequent frames. Since the Kalman filter assumes linear motion dynamics, it may perform poorly when erythrocytes exhibit large motion variations or substantial inter-frame displacements, particularly in arteries and veins where blood flow is relatively fast.

To address this limitation, we propose topology-aware tracking, which incorporates the EMA vessel segmentation mask into the ByteTrack framework to constrain erythrocyte trajectories along anatomically plausible vessel structures. The pseudo-code of the proposed framework is provided in the supplementary materials. Specifically, a Kalman filter is first used to predict the updated location of each track. Following the ByteTrack framework, two rounds of association are performed between the predicted track locations and the detection boxes. For data association, the Intersection-over-Union (IoU) metric is adopted as the similarity measure, with bounding box size of 16 for every detected erythrocyte.

After the two rounds of association, we further perform an additional topology-aware association step using Dijkstra's algorithm \cite{EDijkstraNM1959} to handle unmatched predicted track and detection boxes. The EMA vessel segmentation mask defines the feasible foreground region for path searching. A detection box is associated with a predicted track only if (1) a valid path exists between them within the vessel mask and (2) the path length is smaller than a predefined threshold $D_v=100$. This vessel-constrained association mechanism enables robust matching under large inter-frame displacements while preserving physiologically motion patterns.

\section{Experiments}
\label{sec:exp}
\subsection{Dataset}
The RBF-EMA dataset consists of 58 EMA video clips. For each clip, moving erythrocytes were manually annotated in every frame, along with their corresponding tracking trajectories. Additionally, for each track, the first and last frames in which the erythrocyte is identified within a capillary were annotated to facilitate retinal blood flow (RBF) quantification across different vessel types (\ie, artery, capillary, and vein). An example frame from an EMA video, together with its moving erythrocyte annotations and the corresponding tracking trajectories for the entire video, is provided in the supplementary material.

The EMA videos were acquired from four eyes of two rhesus monkeys (macaca mulatta). All video clips were collected under a protocol approved by the relevant Institutional Animal Care and Use Committee. Detailed experimental procedures are provided in the supplementary material. 

Among the video clips, 19 have a spatial resolution of 512 x 512 pixels, 2 have a resolution of 768 x 768 pixels, and 24 have a resolution of 384 x 384 pixels. During pre-processing, all videos are resized to 512 x 512 pixels, and the annotations are scaled accordingly. For dataset partitioning, 32 video clips are used for training, and the remaining 26 clips are used for testing.

\subsection{Implementation Details and Experimental Setup}
All networks in our method were implemented in PyTorch. FlowContext-DINO was trained using binary cross-entropy (BCE) loss for 50 epochs with a batch size of 2. The Adam optimizer \cite{kingma2014adam} was used with a learning rate of 0.01. The model from the final epoch was selected for evaluation. All training processes were executed on an NVIDIA GeForce RTX 3060 Ti GPU. OpenCV was used to generate the EMA vessel segmentation masks, where \textit{cv2.connectedComponentsWithStats} was applied to extract the centroid of each bright spot. In addition, \textit{cKDTree} and \textit{minimum\_spanning\_tree} from SciPy \cite{scipy_2020} were used to partition all detected points into connected components and to connect nodes within each component.

\subsection{Comparison to Baseline Methods}
For moving erythrocyte detection, we compared FlowContext-DINO with five baseline methods representative of particle detection approaches. Specifically, we included the traditional Laplacian of Gaussian (LoG) method \cite{van2014scikit} as a non–deep learning baseline for white-dot detection. Among deep learning–based methods, deepBlink \cite{eichenberger2021deepblink} has been widely adopted for particle and cell detection tasks and has also been used as the detection module in several particle tracking studies \cite{zhang2023motion}. We evaluated both the pre-trained deepBlink model and a version fine-tuned on our training set.

To further demonstrate the effectiveness of FlowContext-DINO, we additionally evaluated a baseline (i.e., DINO + cosine) that directly utilizes raw DINO features. Specifically, the cosine distance was computed between the feature of each patch in a testing frame and the features of annotated erythrocyte patches from the training set. A predefined similarity threshold (\ie., 0.5) was then applied to identify patches corresponding to moving erythrocytes.

For erythrocyte tracking, we compared topology-aware tracking with ByteTrack \cite{zhang2022bytetrack} to highlight the benefit of incorporating the vessel segmentation mask in the EMA dataset. ByteTrack has also demonstrated strong performance in particle tracking under low-density conditions \cite{zhang2023motion}, which is consistent with the characteristics of our EMA dataset.

\subsection{Evaluation Metrics}
For erythrocyte detection, recall, precision and F1-score are used to evaluate detection performance. For tracking evaluation, we report the total number of missed tracks, total number of spurious tracks, track recall, $\alpha$, $\beta$ and $JSC_\theta$ \cite{chenouard2014objective}. The metric $\alpha \in [0,1]$ measures how well the predicted tracks align with the ground-truth tracks. The metric $\beta \in [0,\alpha]$ is similar to $\alpha$, but additionally penalizes false-positive tracks. $JSC_\theta \in [0,1]$ represents the ratio of matched ground-truth tracks to the total predicted tracks. Except for the total number of missed tracks and the total number of spurious tracks, higher values indicate better performance for all other metrics. Detailed calculation of each metric can be referred to supplementary material.

For RBF quantification, the velocity ($v$) of each track is computed by aggregating the inter-frame displacement (step size) across the entire trajectory and dividing it by the track duration, which can is defined as
\begin{equation}
  v = \frac{1}{T} \sum_{t=1}^{T} d_{t-1,t},
\end{equation}
\begin{equation}
  d_{t-1,t} = \sqrt{(x_t - x_{t-1})^2 + (y_t - y_{t-1})^2},
\end{equation}
where T denotes the total number of frames in the track, and ($x_t$, $y_t$) represents the center coordinates of the detection box at frame t. We report the mean velocity over all matched predicted tracks and their corresponding ground-truth tracks.

\section{Results}
\label{sec:results}
\subsection{Moving Erythrocyte Detection Performance}
\label{sec:results_det}
\begin{table}[tb]
  \caption{Moving cell detection results of different methods on the RBF-EMA dataset. (Best results are marked in \textbf{bold})}
  \label{tab:detection}
  \centering
    \resizebox{0.6\columnwidth}{!}{%
  \begin{tabular}{@{}cccc@{}}
    \toprule
    Method & Recall & Precision & F1 \\
    \midrule
    Traditional Gaussian & 0.497 & 0.063 & 0.112 \\
    DeepBlink (pretrained) & 0.904 & 0.027 & 0.053 \\
    DeepBlink (finetune) & 0.457 & 0.136 & 0.210 \\
    DINO + cosine &  {\bf0.914} & 0.018 & 0.035 \\
    \midrule
    {\bf FlowContext-DINO (Ours)} & 0.649 & {\bf 0.576} & {\bf 0.611} \\
   \bottomrule
  \end{tabular}
  }
\end{table}

\begin{figure}[tb]
  \centering
  \begin{subfigure}{0.27\linewidth}
  \includegraphics[width=\linewidth]{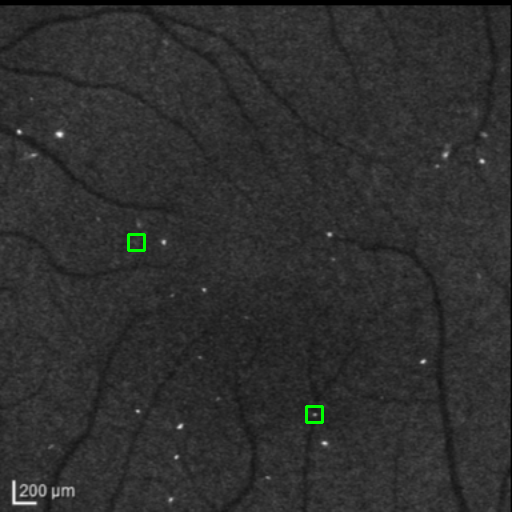}
    \caption{Ground Truth}
    \label{fig:results1-a}
  \end{subfigure}
  \hfill
  \begin{subfigure}{0.27\linewidth}
    \includegraphics[width=\linewidth]{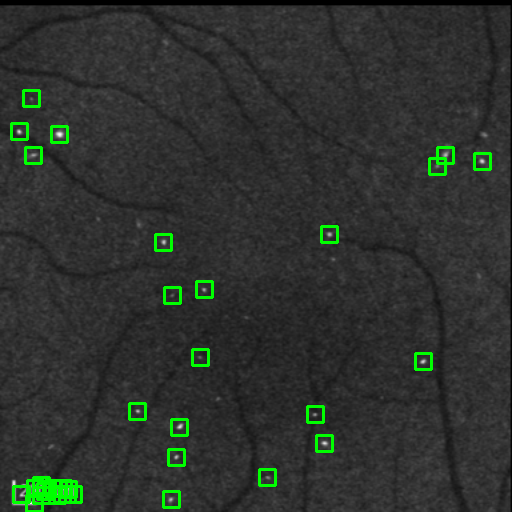}
    \caption{Traditional Gaussian}
    \label{fig:results1-b}
  \end{subfigure}
  \hfill
  \begin{subfigure}{0.27\linewidth}
    \includegraphics[width=\linewidth]{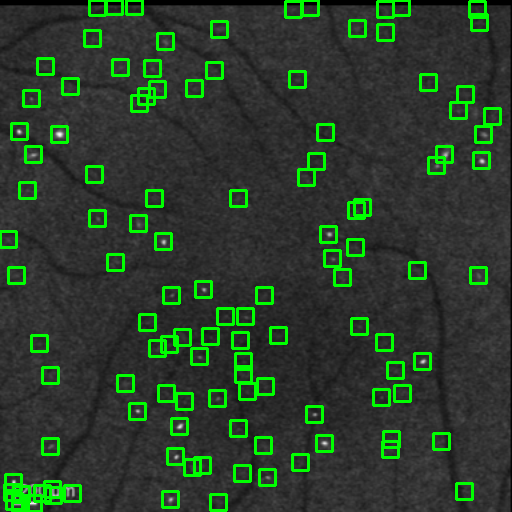}
    \caption{Deepblink (pretrained)}
    \label{fig:results1-c}
  \end{subfigure}
  
  \vspace{0.3cm}
  
  \begin{subfigure}{0.27\linewidth}
    \includegraphics[width=\linewidth]{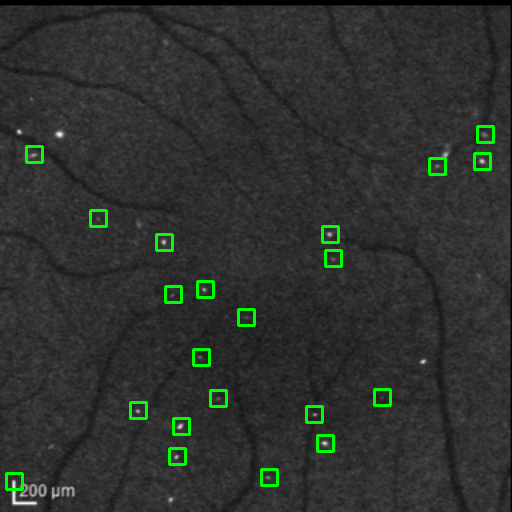}
    \caption{Deepblink (finetune)}
    \label{fig:results1-d}
  \end{subfigure}
  \hfill
  \begin{subfigure}{0.27\linewidth}
    \includegraphics[width=\linewidth]{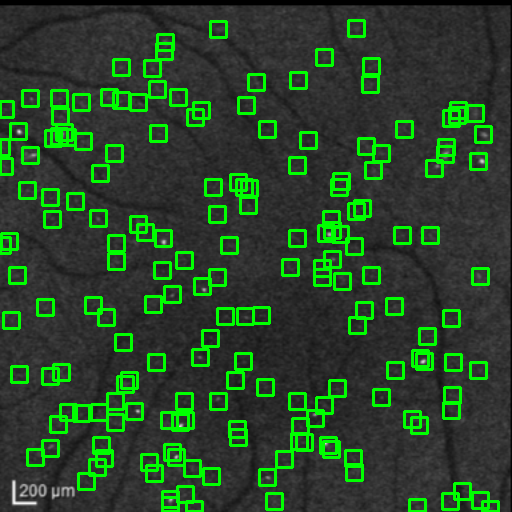}
    \caption{DINO+cosine}
    \label{fig:results1-e}
  \end{subfigure}
  \hfill
  \begin{subfigure}{0.27\linewidth}
    \includegraphics[width=\linewidth]{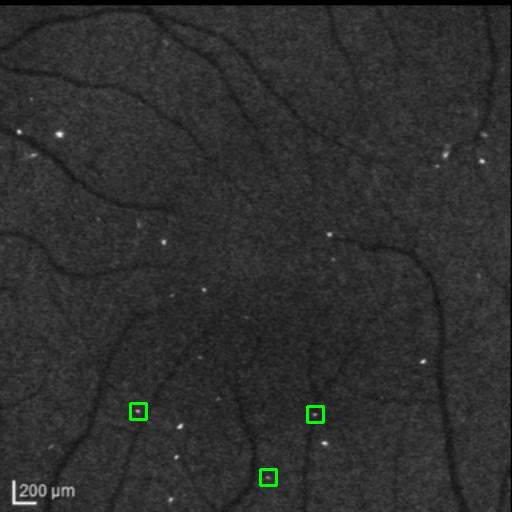}
    \caption{FlowContext-DINO}
    \label{fig:results1-f}
  \end{subfigure}
  \caption{Qualitative comparison of different detection methods. Predicted cells are indicated by green boxes.}
  \label{fig:results1}
\end{figure}

\cref{tab:detection} presents the quantitative results of moving erythrocyte detection, comparing our method with the baseline approaches. Qualitative results on a representative image from the RBF-EMA dataset are shown in \cref{fig:results1}. Generally, our method outperforms all baseline methods in terms of precision and F1-score. 

From \cref{tab:detection} and \cref{fig:results1}, it can be observed that the Traditional-Gaussian method performs poorly on the RBF-EMA dataset, as it fails to distinguish moving cells from stationary cells and background noise. This suggests that conventional methods are insufficient to capture the complex visual characteristics present in this scenario.

The two deep-learning based method, DINO-cosine and DeepBlink, achieve high recall but relatively low precision. This indicates that, although they successfully detect most moving cells, they also produce a substantial number of false positives by misclassifying stationary cells and background noise as moving cells. While DINO features contain rich semantic and structural information, single-frame representations are insufficient to reliably distinguish moving cells from stationary ones. 

In contrast, FlowContext-DINO is the only method that effectively differentiates moving cells from stationary cells (\cref{fig:results1}), even when their visual appearances are highly similar. These results demonstrate that temporal information is critical for accurate moving-cell detection, and that our model architecture effectively integrates temporal cues with the rich semantic representations extracted from DINO features.

\subsection{Tracking Performance}

\begin{table}[tb]
  \caption{Quantitative comparison of ByteTrack and our topology-aware tracking method across different detection methods. For each metric, the left value indicates the result obtained using the standard ByteTrack tracker, while the right value shows the result after applying topology-aware tracking. The value in parentheses (highlighted in green) denotes the improvement achieved by the proposed method.}
  \label{tab:tracking}
  \centering
  \resizebox{0.8\columnwidth}{!}{%
  \setlength{\tabcolsep}{8pt}
  \begin{tabular}{@{}cccc@{}}
    \toprule
    Detection Method & $\alpha$ & $\beta$ & $JSC_\theta$ \\
    \midrule
    Ground Truth Labeling &  \makecell{0.481 $\rightarrow$ 0.695 \\ (\textcolor{mildgreen}{+0.214})} & \makecell{0.471 $\rightarrow$ 0.672 \\ (\textcolor{mildgreen}{+0.201})} & \makecell{0.830 $\rightarrow$ 0.859 \\ (\textcolor{mildgreen}{+0.029})}\\
    Traditional Gaussian &  \makecell{0.181 $\rightarrow$ 0.272\\ (\textcolor{mildgreen}{+0.091})}& \makecell{0.017 $\rightarrow$ 0.023 \\ (\textcolor{mildgreen}{+0.006})} &  \makecell{0.070 $\rightarrow$ 0.071 \\ (\textcolor{mildgreen}{+0.001})} \\
    DeepBlink (finetune)  & \makecell{0.154 $\rightarrow$ 0.223 \\ (\textcolor{mildgreen}{+0.069})} & \makecell{0.031 $\rightarrow$ 0.043 \\ (\textcolor{mildgreen}{+0.012})} & \makecell{0.125 $\rightarrow$ 0.134 \\ (\textcolor{mildgreen}{+0.009})} \\
    \midrule
    \bf{FlowContext-DINO (ours)}  & \makecell{0.219 $\rightarrow$ 0.325 \\ (\textcolor{mildgreen}{+0.106})} & \makecell{0.172 $\rightarrow$ 0.248 \\ (\textcolor{mildgreen}{+0.076})}  & \makecell{0.436 $\rightarrow$ 0.463\\ (\textcolor{mildgreen}{+ 0.027})} \\
    
   \bottomrule
  \end{tabular}
}
\end{table}

\begin{figure}[ht!]
  \centering
  \includegraphics[width=0.9\linewidth]{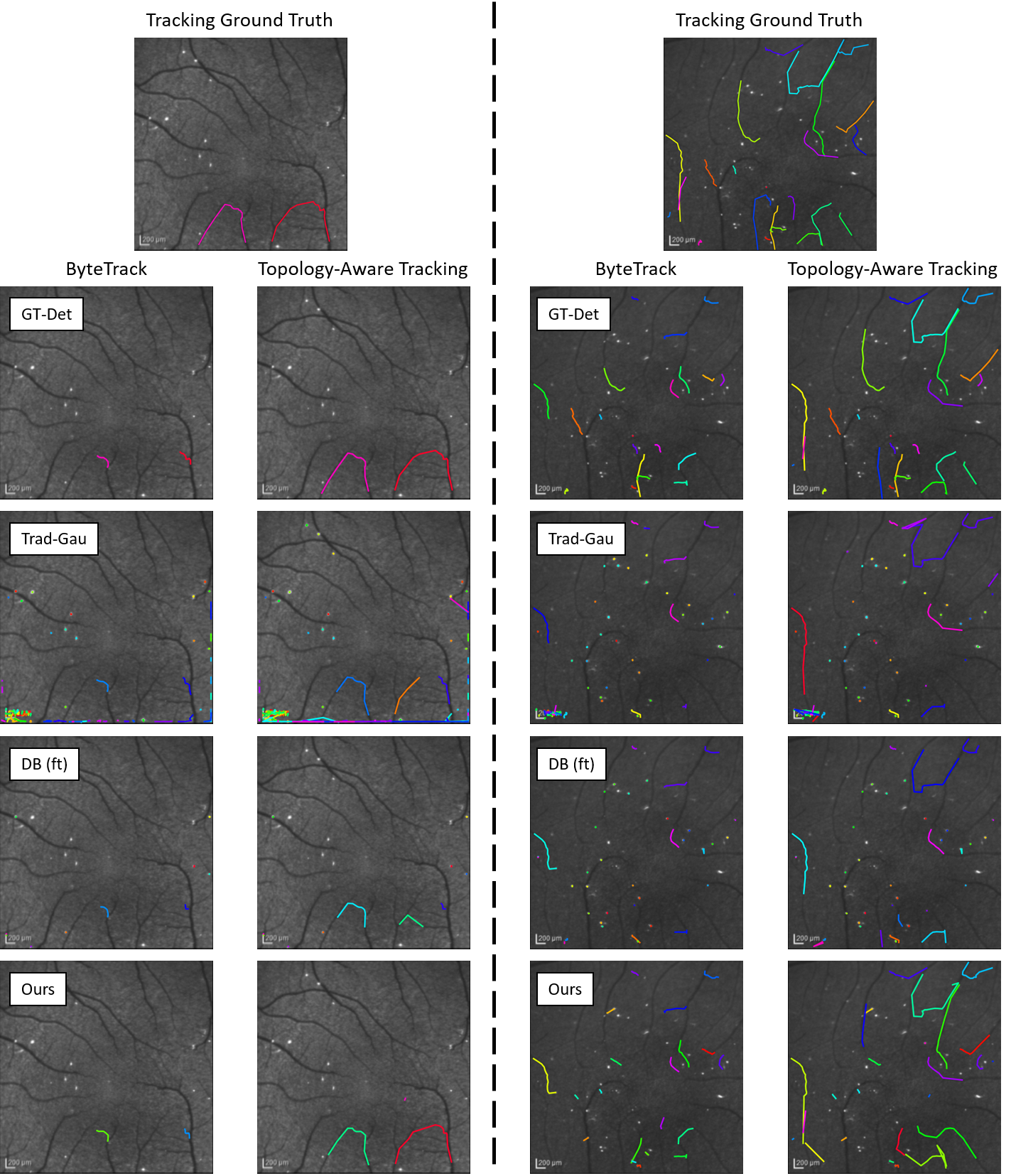}
  \caption{Qualitative comparison of tracking results between ByteTrack and our proposed topology-aware tracking method using different detection methods. GT-Det denotes the use of ground-truth detections; Trad-Gau denotes the traditional Gaussian-based detector; DB (ft) denotes deepBlink fine-tuned on our training set.}
  \label{fig:results2}
\end{figure}

To demonstrate the effectiveness of topology-aware tracking in the RBF-EMA dataset, we compare its tracking performance that of the original ByteTrack across three detection methods (i.e. traditional Gaussian, Deepblink (finetune) and Ours) in \cref{tab:tracking}. In addition, tracking performance using ground-truth detections is reported to enable a controlled comparison. Representative qualitative results on the RBF-EMA dataset are also provided in \cref{fig:results2} to further illustrate the differences between the two approaches.

The results show that topology-aware tracking consistently outperforms ByteTrack across all evaluation metrics and achieves superior performance regardless of the detection method used, including when evaluated with ground-truth detections. Furthermore, ByteTrack exhibits missed tracks in large vessels (i.e., arteries and veins) and frequent ID switches when cells move from large vessels into capillaries. These observations indicate that topology-aware tracking more effectively associates detections in scenarios characterized by large inter-frame displacements and substantial motion variations. 

Regarding tracking performance under different detection methods, the combination of FlowContext-DINO and topology-aware tracking achieves the best results across all metrics. As discussed in Section~\ref{sec:results_det}, the baseline detection methods exhibit relatively low precision and tend to misclassify stationary cells and background noise as moving cells. Such false positives significantly degrade tracking performance by generating spurious tracks and incorrect data associations between erroneous detections \cref{fig:results2}). Moreover, tracking performance using ground-truth detections is superior, suggesting that improvements in detection accuracy directly contribute to better tracking results.

\subsection{Retinal Blood Flow Quantification}
\begin{figure}[tb]
  \centering
  \includegraphics[width= 0.7 \linewidth]{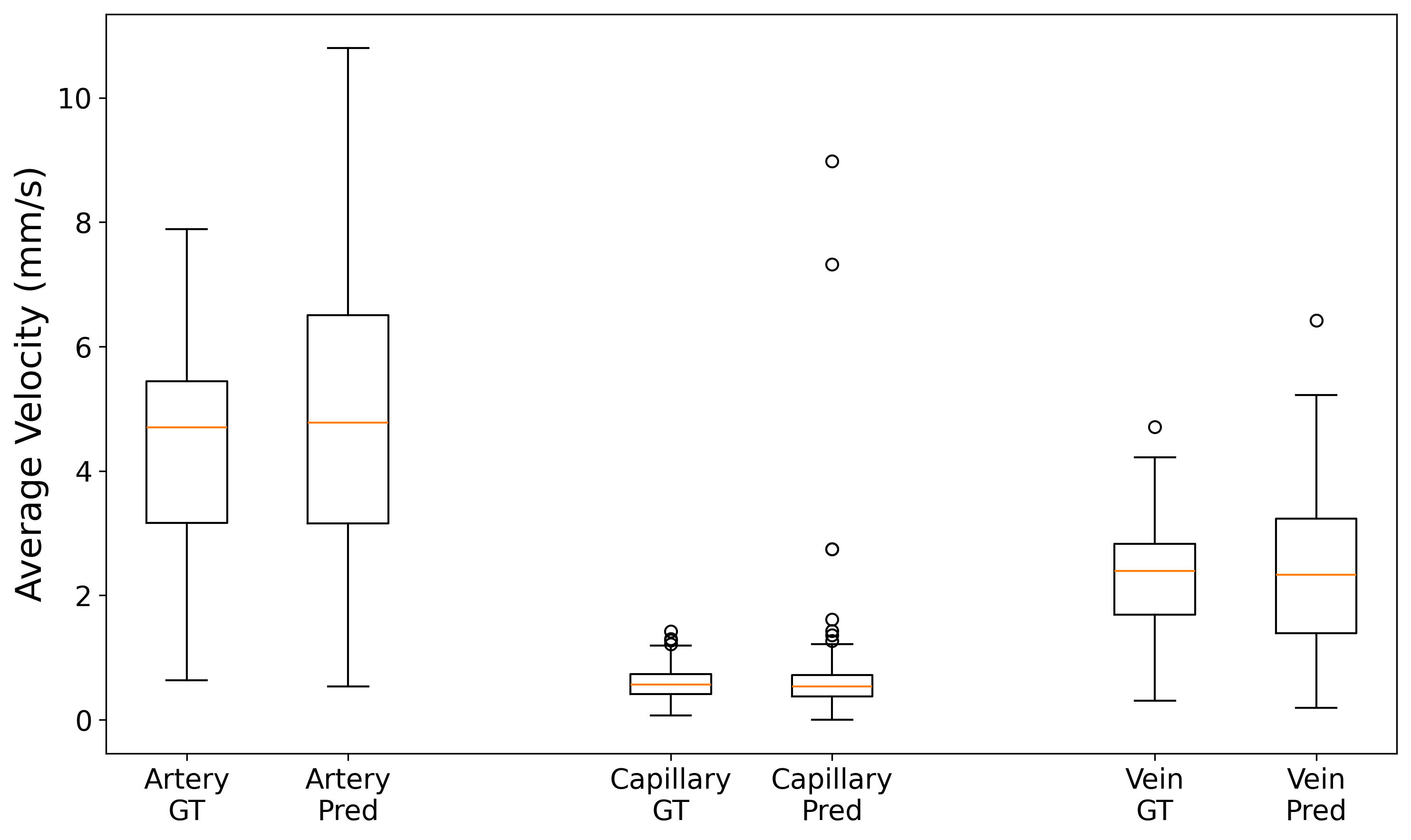}
  \caption{Average velocities of matched predicted and ground-truth tracks across different vessel types. GT denotes ground truth, and Pred denotes predictions.}
  \label{fig:results3}
\end{figure}

To demonstrate the feasibility of using EMTrack for automated retinal blood flow (RBF) quantification, we report the mean velocity computed from all matched predicted tracks generated by our method and their corresponding ground-truth tracks in \cref{fig:results3} across different vessel types (i.e., arteries, capillaries, and veins). 

As shown in the figure, the median velocities of the predictions are close to those of the ground truth across all vessel types. Moreover, the overall distributions of the predicted average velocities are largely consistent with the ground-truth distributions. However, the predicted velocity distribution for arteries appears slightly wider than that of the ground truth. This discrepancy may be attributed to the larger cell displacements typically observed in arteries, as well as the relatively smaller number of tracked cells in arteries compared with other vessel types due to their higher flow speeds. In addition, several outliers are observed in the predicted capillary velocities. A possible explanation is tracking noise caused by occasional missed detections within a trajectory, which can affect the velocity estimation. Overall, these results suggest that our framework has strong potential for reliable and automated RBF quantification.

\section{Conclusion}
\label{sec:conclusion}
In this paper, we propose an automated framework for retinal blood flow quantification, termed EMTrack. We design a lightweight temporal module that learns residual features from DINO features for moving cell detection (FlowContext-DINO), and we introduce a label-free cell tracking module that leverages vessel segmentation masks (topology-aware tracking). 
In addition, we present RBF-EMA dataset with both detection and tracking annotations for comprehensive evaluation. Compared to existing cell or particle tracking datasets, this dataset highlights unique challenges, including the presence of stationary cells, large inter-frame displacements, and substantial motion variations. 
Both quantitative and qualitative results demonstrate that our method outperforms baseline approaches. Specifically, the proposed framework effectively detects moving cells and addresses the unique tracking challenges in EMA data. Furthermore, the RBF quantification results indicate the strong potential of our method for automated retinal blood flow measurement. 
We hope that this work will broaden the utilization of EMA for RBF quantification and accelerate the translation of retinal blood flow into a reliable clinical biomarker.

%
%
\bibliographystyle{splncs04}
\bibliography{main}
\end{document}